\documentclass[10pt,twocolumn,letterpaper]{article}

\usepackage{cvpr}
\usepackage{times}
\usepackage{epsfig}
\usepackage{graphicx}
\usepackage{amsmath}
\usepackage{amssymb}
\usepackage{multirow}
\usepackage{enumitem}
\usepackage{subcaption}
\usepackage{xspace}
\usepackage{placeins}
\usepackage{floatrow}
\usepackage{siunitx}
\usepackage{url}
\usepackage[dvipsnames]{xcolor}
\graphicspath{{./figures/}}

\usepackage[pagebackref=true,breaklinks=true,letterpaper=true,colorlinks,bookmarks=false]{hyperref}
\usepackage{cleveref}

\cvprfinalcopy % *** Uncomment this line for the final submission

\def\cvprPaperID{2138} % *** Enter the CVPR Paper ID here

\ifcvprfinal\fi
\begin{document}
\newcommand{\dsetName}{ContactDB\xspace}

%%%%%%%%% TITLE
\title{\dsetName: Analyzing and Predicting Grasp Contact via Thermal Imaging}

\author{
Samarth Brahmbhatt$^1$, Cusuh Ham$^1$, Charles C. Kemp$^1$ and James Hays$^{1,2}$\\
$^1$Institute for Robotics and Intelligent Machines, Georgia Tech $^2$Argo AI\\
%{\tt\small \{samarth.robo, cusuh, charlie.kemp, hays\}@gatech.edu}
{\small \url{{samarth.robo, cusuh}@gatech.edu}, \url{charlie.kemp@bme.gatech.edu}, \url{hays@gatech.edu}}
}

\twocolumn[{%
\renewcommand\twocolumn[1][]{#1}%
\maketitle
\centering
\includegraphics[width=.95\textwidth]{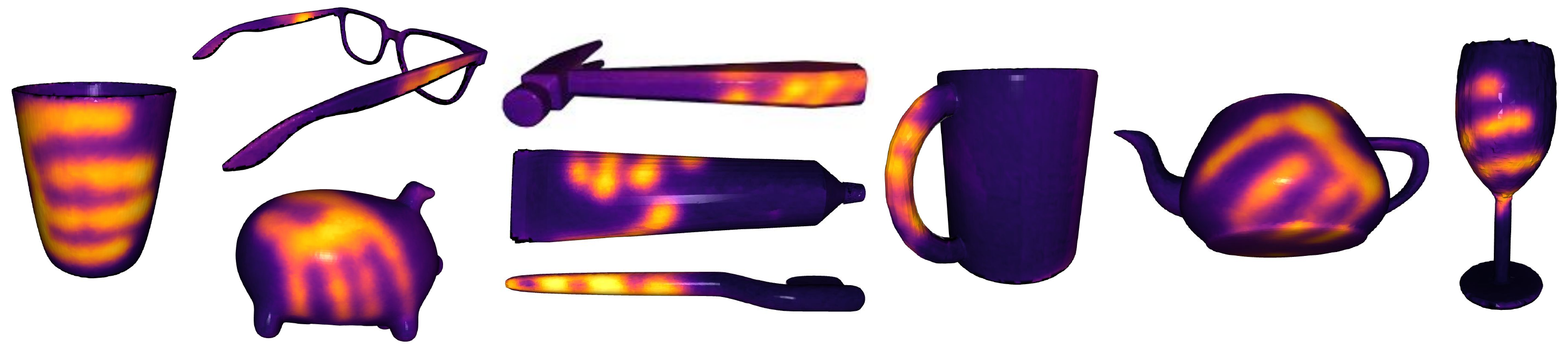}
\captionof{figure}{Example contact maps from \dsetName, constructed from multiple 2D thermal images of hand-object contact resulting from human grasps.}
\label{fig:teaser}
}]

\maketitle
%\thispagestyle{empty}

%%%%%%%%% ABSTRACT
\begin{abstract}
Grasping and manipulating objects is an important human skill. Since hand-object contact is fundamental to grasping, capturing it can lead to important insights. However, observing contact through external sensors is challenging because of occlusion and the complexity of the human hand. We present \dsetName,  a novel dataset of contact maps for household objects that captures the rich hand-object contact that occurs during grasping, enabled by use of a thermal camera. Participants in our study grasped 3D printed objects with a post-grasp functional intent. \dsetName includes 3750 3D meshes of 50 household objects textured with contact maps and 375K frames of synchronized RGB-D+thermal images. To the best of our knowledge, this is the first large-scale dataset that records detailed contact maps for human grasps. Analysis of this data shows the influence of functional intent and object size on grasping, the tendency to touch/avoid `active areas', and the high frequency of palm and proximal finger contact. Finally, we train state-of-the-art image translation and 3D convolution algorithms to predict diverse contact patterns from object shape. Data, code and models are available at \url{https://contactdb.cc.gatech.edu}.
\end{abstract}

\section{Introduction} \label{sec:intro}
Humans excel at grasping and then performing tasks with household objects. Human grasps exhibit contact locations, forces and stability that allows post-grasp actions with objects, and are also significantly influenced by the post-grasp intent~\cite{grasp_influence, intent_influence0, intent_influence1}. For example, people typically grasp a knife by the handle to use it, but grasp it by the blunt side of the blade to hand it off.

A large body of previous work~\cite{glove_ann_grasp0, glove_ann_grasp1, tax_ann_grasp0, tax_ann_grasp1, manual_ann_grasp0, manual_ann_grasp1, passive_ann_grasp, vision_grasp_tax_pred1, vision_grasp_tax_pred2, tax_ann_grasp0, vision_grasp_tax_pred2, tax_ann_grasp2, tax_ann_grasp1} has recorded human grasps, with methods ranging from data gloves that measure joint configuration to manually arranged robotic hands. \dsetName differs significantly from these previous datasets by \textit{focusing primarily on the contact} resulting from the rich interaction between hand and object. Specifically, we represent contact through the texture of 3D object meshes, which we call `contact maps' (see Figure~\ref{fig:teaser}).

There are multiple motivations for recording grasping activity through contact maps. Since it is \textit{object-centric}, it enables detailed analysis of grasping preferences influenced by functional intent, object shape, size and semantic category, and learning object shape features for grasp prediction, and grasp re-targeting to kinematically diverse hand models. Previously employed methods of recording grasping activity do not easily support such analysis, as we discuss in Section~\ref{sec:related_work}.

We created \dsetName by recording human participants grasping a set of 3D printed household objects in our laboratory, with two different post-grasp functional intents--using the object and handing it off. See Section~\ref{sec:dataset} for more details on the data collection procedure, size of the dataset and the kinds of data included.

Except for contact edges viewed from select angles, and contact with transparent objects, contact regions are typically occluded from visual light imaging. Hence, existing studies on the capture and analysis of hand-object contact are extremely limited. Fundamental questions such as the  role of the palm in grasping everyday objects are unanswered. We propose a novel procedure to capture contact maps on the object surface at unprecedented detail using an RGB-D + thermal camera calibrated rig.

We make the following contributions in this paper:
\begin{itemize}[noitemsep,topsep=0pt,leftmargin=*]
	\item \textbf{Dataset}: Present a dataset recording functional human grasping consisting of 3750 meshes textured with contact maps and 375K frames of paired RGBD-thermal 
	data.
	\item \textbf{Analysis}: Demonstrate the influence of object shape, size and functional intent on grasps, and show the importance of non-fingertip contact.
	\item \textbf{Prediction}: Explore data representations and diverse prediction algorithms to predict contact maps from object shape.
\end{itemize}
\section{Related Work} \label{sec:related_work}
\subsection{Datasets of Human Grasps}
Since contact between the human hand and an object is fundamental to grasping and manipulation, capturing this contact can potentially lead to important insights about human grasping and manipulation. In practice, however, this has been a challenging goal. The human hand is highly complex with extensive soft tissue and a skeletal structure that is often modeled with 26 degrees of freedom. Hence, previous work has focused on recording grasping activity in other forms like hand joint configuration by manual annotation~\cite{manual_ann_grasp0, manual_ann_grasp1}, data gloves~\cite{glove_ann_grasp0, glove_ann_grasp1} or wired magnetic trackers~\cite{yuan2017bighand2, garcia2018first} (which can interfere with natural grasping), or model-based hand pose estimation~\cite{passive_ann_grasp}. At a higher level, grasping has been observed through third-person~\cite{vision_grasp_tax_pred1, vision_grasp_tax_pred2, tax_ann_grasp0} or first-person~\cite{vision_grasp_tax_pred2, tax_ann_grasp2, tax_ann_grasp1} videos, in which frames are annotated with the category of grasp according to a grasp taxonomy~\cite{grasp_taxonomy0, grasp_taxonomy1}. Tactile sensors are embedded on a glove~\cite{glove_tactile_sensors} or in the object~\cite{pham2018hand} to record grasp contact points. Such methods are limited by the resolution of tactile sensors. Puhlmann et al~\cite{puhlmann2016compact} capture hand-table contact during grasping with a touchscreen. Rogez et al~\cite{rogez2015understanding} manually configure a hand model to match grasps from a taxonomy, and use connected component analysis on hand vertices intersecting with an object model to estimate contact regions on the hand.

Due to hand complexity and lack of understanding of how humans control their hands, approaches like those mentioned above have so far been limited to providing coarse or speculative contact estimates. In contrast, our approach allows us to directly observe where contact between the object and the human hand has taken place with an unprecedented level of fidelity.

\subsection{Predicting Grasp Contact}
Our work is related to that of Lau et al~\cite{tactile_mesh_saliency}, which crowdsources grasp tactile saliency. Online annotators are instructed to choose a point they would prefer to touch, from a pair sampled from the object surface. This pairwise information is integrated to construct the tactile saliency map. In contrast, \dsetName contact maps are full observations of real human grasps with functional intent (see supplementary material for a qualitative comparison). Akizuki et al~\cite{tactile_logging} use hand pose estimation and model-based object tracking in RGB-D videos to record a set of contact points on the object surface. This is vulnerable to inaccuracies in the hand model and hand pose tracking. Hamer at al~\cite{hamer2010object} record human demonstrations of grasping by registering depth images to get object geometry and object- and hand-pose. Contact is approximated as a single point per fingertip. A large body of work in robotics aims to predict a configuration of the end-effector~\cite{dexnet, choi_grasping, saxena_grasping} suitable for grasping. In contrast to \dsetName, these works model contact as a single point per hand digit, ignoring other contact.

\textbf{Diverse Predictions}: Grasping is a task where multiple predictions can be equally correct. Lee et al~\cite{smcl} and Firman et al~\cite{diversenet} have developed theoretical frameworks allowing neural networks to make diverse and meaningful predictions. Recently, Ghazaei et al~\cite{ghazaei2018dealing} have used similar techniques to predict diverse grasp configurations for a parallel jaw gripper.
\section{The \dsetName Dataset} \label{sec:dataset}

\begin{figure*}
\begin{subfigure}{.28\textwidth}
	\includegraphics[width=\textwidth]{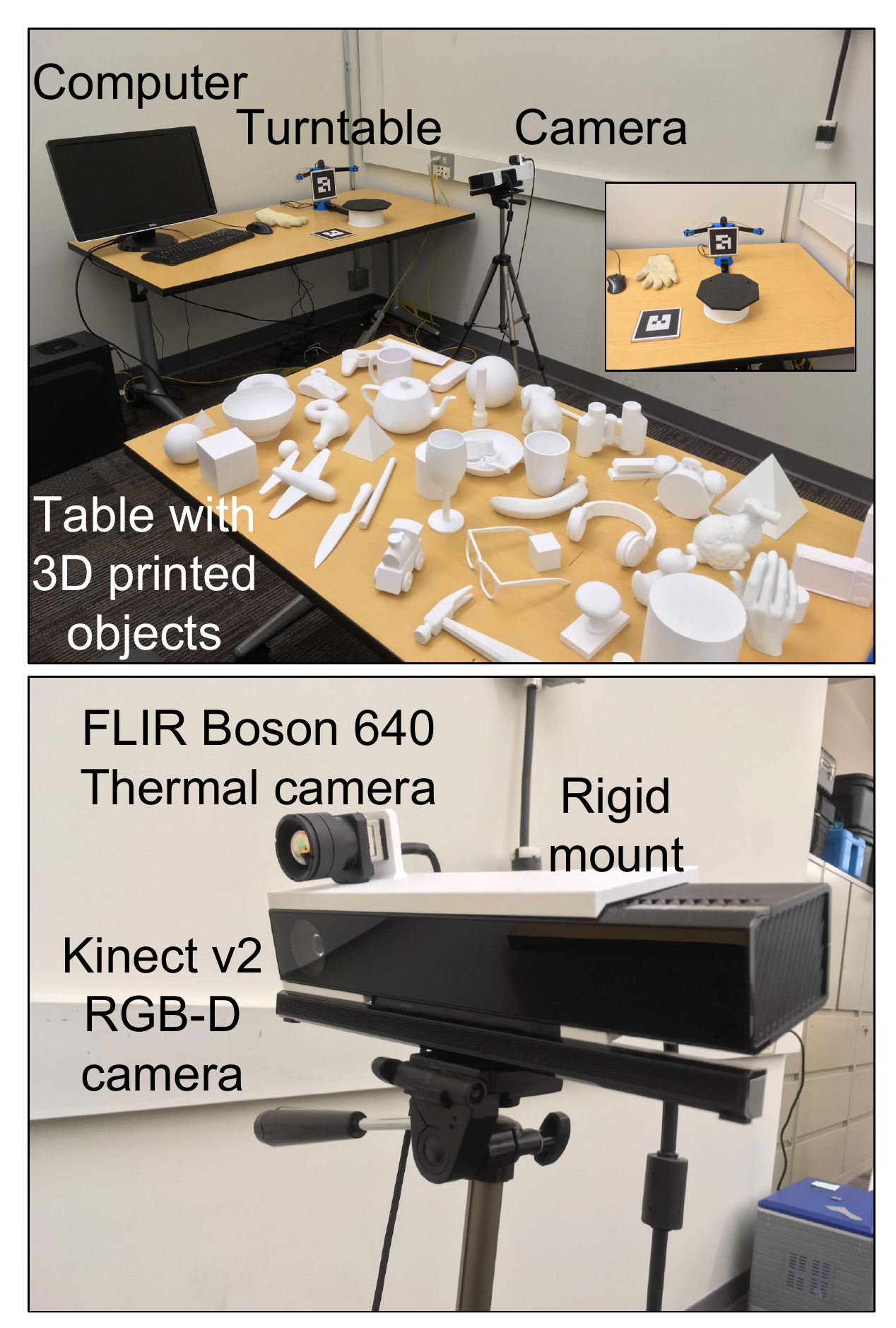}
	\caption{Data collection area setup}
	\label{fig:data_collection}
\end{subfigure}
\hfill
\begin{subfigure}{.7\textwidth}
	\includegraphics[width=\textwidth]{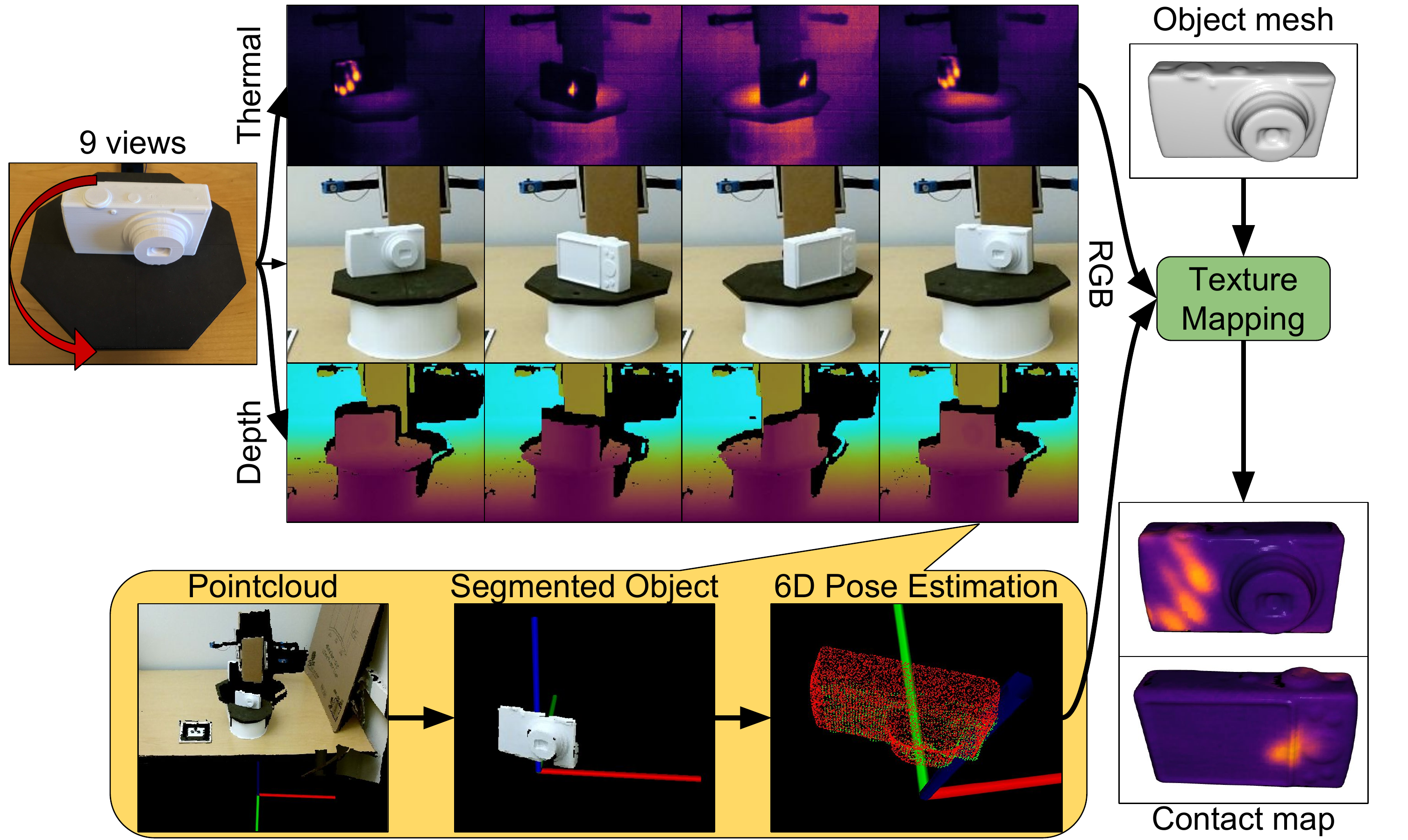}
	\caption{Data processing pipeline, explained in detail in Section~\ref{subsec:data_processing}}
	\label{fig:data_processing}
\end{subfigure}
\caption{Data collection and processing for \dsetName. Participants grasp 3D printed objects and put them on the rotating turntable. Thermal images from multiple views are 
texture-mapped to the object mesh.}
\end{figure*}

Here we present the design choices and process in creating the \dsetName, which consists of 50 3D printed household objects being grasped with two functional
intents by 50 participants (see Table~\ref{tab:dataset_size}).

\textbf{Observing Contact Through a Thermal Camera.}
At the core of our data collection process is the use of a thermal camera to observe the precise locations of contact between human hand and object. Thermal cameras have recently been used to capture humans and their interaction with the environment. For example, Luo et al~\cite{luo2017scene} observe humans interacting with objects for egocentric SLAM, while Larson et al~\cite{Larson:2011:HTI:1978942.1979317} observe human finger interaction with arbitrary surfaces to make them interactive. Both note the phenomenon of thermally observable contact, but do not investigate it rigorously or collect a large-scale dataset.

When a participant grasps an object, heat from the hand transfers onto the object surface. If the object material does not dissipate the heat rapidly, the precise contact areas can be clearly observed in the thermal image after the object is released (see Figure~\ref{fig:data_processing}). Intensity at a pixel in the thermal image is a function of the infrared energy emitted by the corresponding world point~\cite{vollmer2017infrared}. Hence, object pixel intensity in our thermal images is related to heat of the skin, duration of contact, heat conduction (including diffusion to nearby object locations), and contact pressure. By keeping these factors roughly constant during data collection, we verified empirically that heat conduction from hand-object contact is the dominant factor in the observed thermal measurements. See the supplementary material for more discussion on heat dissipation and accuracy.

\begin{table}
\centering
\small
\begin{tabular}{cc|c|c}
\multirow{2}{*}{} & \multicolumn{2}{c}{\textbf{Functional Intent}} & \multirow{2}{*}{\textbf{Total}} \\
& \textbf{Use} & \textbf{Hand-off} &\\
\hline
Participants & 50 & 50 (same) &\\
Objects & 27 & 48 (overlapping) & 50\\
Textured meshes & 1350 & 2400 & \textbf{3750} \\
RGBD-Thermal frames & 135K & 240K & \textbf{375K} \\
\end{tabular}
\caption{Size of the \dsetName Dataset}
\label{tab:dataset_size}
\end{table}

\subsection{Object Selection and Fabrication}
We decided to focus on household objects since an understanding of contact preferences and the ability to predict them are most likely to improve human-robot interaction in household settings. Other standard grasping datasets~\cite{ycb} and competitions~\cite{amazon_picking_challenge} have a similar focus. We started with the YCB dataset~\cite{ycb} to choose the 50 objects in our dataset. We excluded similarly-shaped objects (e.g. cereal and cracker boxes) that are unlikely to produce different kinds of grasps, deformable objects (e.g. sponge, plastic chain, nylon rope), very small (e.g. dominoes, washers), and very large objects (e.g. cooking skillet, Windex bottle). We added common ones such as flashlight, eyeglasses, computer mouse, and objects popular in computer graphics (e.g. Stanford bunny and Utah teapot). Since object size has been shown to influence the grasp~\cite{size_influence, grasp_influence} and we are interested in contact during grasping of abstract shapes, we included 5 primitive objects--cube, cylinder, pyramid, torus and sphere--at 3 different scales (principal axes 12, 8 and 4 cm). See the supplementary material for a full object list.

We chose to 3D print all the objects to ensure
uniform heat dissipation properties. Additionally, we empirically found that the PLA material used for 3D printing is excellent for retaining thermal handprints.
We used open-source resources to select suitable models for each object, and printed them at 15\% infill density using white PLA filament on a Dremel 3D20 printer.
3D printing the objects has additional advantages. Having an accurate 3D model of the object makes
6D pose estimation of the object from recorded pointcloud data easier (see Section~\ref{subsec:data_processing}), which we use for texture mapping contact maps to the
object mesh. 3D printing the objects also allows participants to focus on the object geometry during grasping.

\subsection{Data Collection Protocol}
Figure~\ref{fig:data_collection} shows our setup. We rigidly mounted a FLIR Boson 640 thermal camera on a Kinect v2 RGB-D sensor. The instrinsics of both the cameras and
extrinsics between them are calibrated using ROS~\cite{ros}, so that both RGB and depth images from the Kinect can be accurately registered to the thermal image. We invited 50 participants (mostly 20-25 years of age, able-bodied males and females), and used the following protocol approved by the Georgia Tech Institutional Review Board.

50 3D printed objects were placed at random locations on a table in orientations commonly encountered in practice.
Participants were asked to grasp each object with a post-grasp functional intent. They held the object for 5 seconds
to allow heat transfer from the hand to the object, and then hand it to an experimenter. The experimenter wore an insulating glove to prevent heat transfer from their hand,
and places the object on a turntable about 1 m away from the cameras. Participants were provided with chemical hand warmers to increase the intensity of thermal handprints.
The cameras recorded a continuous stream of RGB, depth and thermal images as the turntable rotated in a 360 degree arc. The turntable
paused at 9 equally spaced locations on this arc, where the rotation angle of the turntable was also recorded. In some cases,
objects were flipped and scanned a second time to capture any thermal prints that were unseen in the previous rotation.

We used two post-grasp \textit{functional intents}: `use' and `hand-off'. Participants were instructed to grasp 48 objects with the intent of handing them off to the experimenter, and to grasp a subset of 27 objects (after the previous thermal handprints had dissipated) with the intent of using them. We used only a subset of 27 objects for `use', since other objects (e.g. pyramid, Stanford bunny) lack clear use cases. See the supplementary material for specific use instructions. Participants were asked to avoid in-hand manipulation after grasping to avoid smudging the thermal handprints.

\subsection{Data Processing} \label{subsec:data_processing}
As the turntable rotates with the object on it, the stream of RGB-D and thermal images capture the object from multiple viewpoints. The aim of data processing
is to texture-map the thermal images to the object 3D mesh and generate a coherent contact map (examples are shown in Figure~\ref{fig:teaser}).

The entire process is shown in Figure~\ref{fig:data_processing}.
We first extracted the corresponding turntable angle and RGB, depth and thermal images at the 9 locations where the turntable pauses. Next, we converted the depth maps to pointclouds and useed a least-squares estimate of the turntable plane and white color segmentation to segment the object. We used the Iterative Closest Point (ICP)~\cite{icp} algorithm implemented in PCL~\cite{pcl} to estimate the full 6D pose of the object in the 9 segmented pointclouds. Object origins in the 9 views were used to get a least squares estimate of the 3D circle described by the moving object. This circle was used to interpolate the object poses for views which are unsuitable for the ICP step because of noise in the depth map or important shape elements of the object being hidden in that view, or for rotating symmetric objects around the axis of symmetry.

Finally, the 3D mesh along with the 9 pose estimates and thermal images were input to the colormap optimization algorithm of~\cite{colormap_optim},
which is implemented in Open3D~\cite{open3d}. It locally optimizes object poses to minimize the photometric texture projection error and generates a mesh
coherently textured with contact maps.
\section{Analysis of Contact Maps} \label{sec:analysis}
In this section we present analysis of some aspects of human grasping, using the data in \dsetName. We processed each contact map separately to increase contrast
by applying a sigmoid function to the texture-mapped intensity values that maps the minimum to 0.05 and maximum to 0.95.

\begin{figure*}[h!]
\centering
\includegraphics[width=0.95\textwidth]{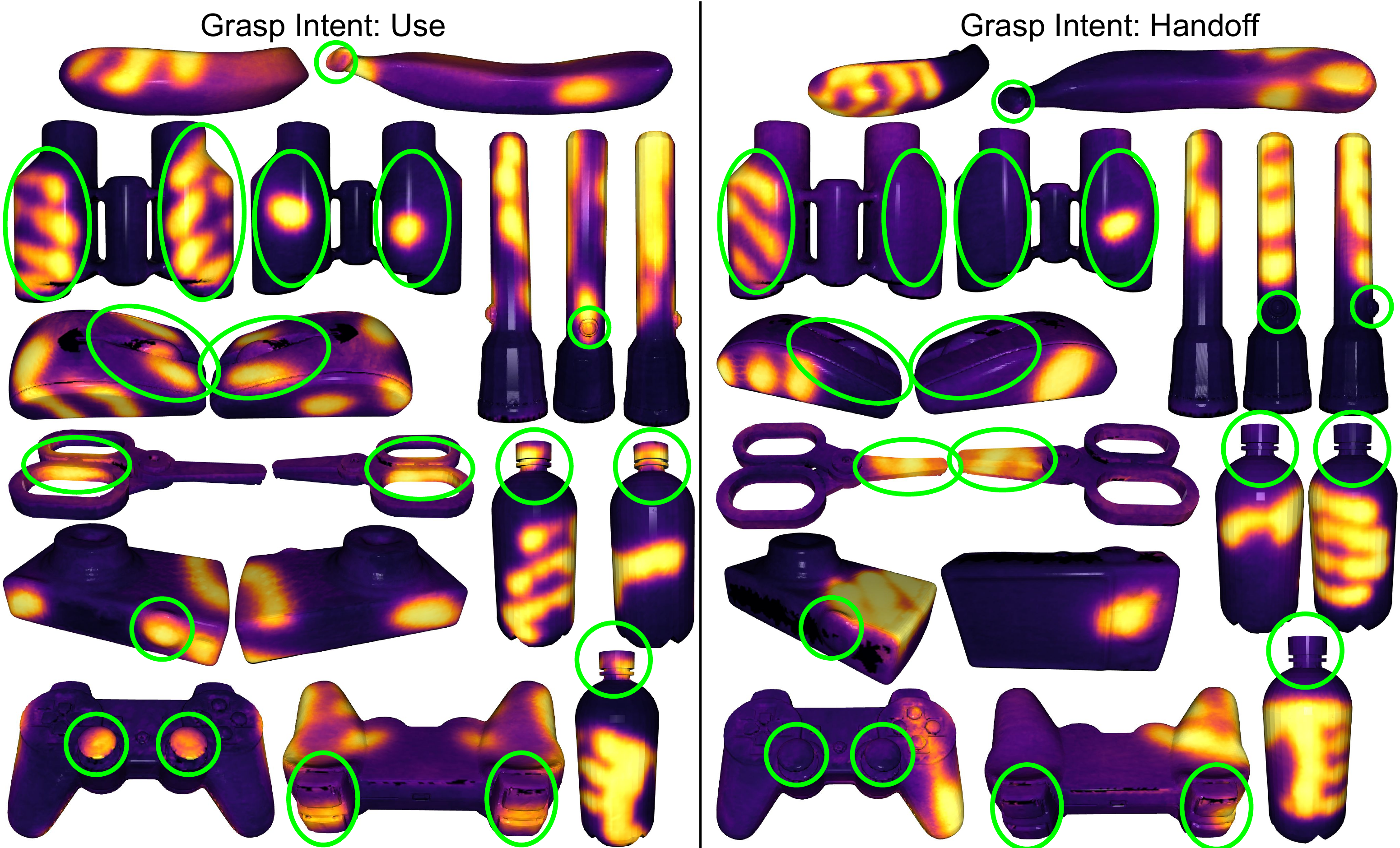}
\caption{Influence of functional intent on contact: Two views of the dominant grasp (center of the largest cluster after $k$-medoids clustering across participants).
\textcolor{OliveGreen}{Green circles} indicate `active areas'. This influence is quantified in Table~\ref{tab:intent_influence}.}
\label{fig:intent_influence}
\end{figure*}
\noindent
\textbf{Effect of Functional Intent.}
We observed that the functional intent (`use' or `hand off') significantly influences the contact patterns for many objects. To show qualitative examples, we clustered the contact
maps within each object and functional intent category using $k$-medoids clustering~\cite{k_medoids} ($k=3$) on the XYZ values of points which have contact value above 0.4.
The distance function between two sets of points was defined as
$d(\mathbf{p}_1, \mathbf{p}_2) = \left(\bar{d}(\mathbf{p}_1, \mathbf{p}_2) + \bar{d}(\mathbf{p}_2, \mathbf{p}_1) \right) / \left(|\mathbf{p}_1| + |\mathbf{p}_2| \right)$, where
$\bar{d}(\mathbf{p}_1, \mathbf{p}_2) = \sum_{i=1}^{|\mathbf{p}_1|} \min_{j=1}^{|\mathbf{p}_2|} ||\mathbf{p}_1^{(i)} - \mathbf{p}_2^{(j)}||_2$.
For symmetric objects, we chose the angle of rotation around the axis of symmetry that minimized $d(\mathbf{p}_1, \mathbf{p}_2)$. Figure~\ref{fig:intent_influence} 
shows dominant contact maps (center of the largest cluster) for the two different functional intents.

\begin{table}
\centering
\small
\begin{tabular}{c|c|c}
\textbf{Active Area}                        & \textbf{handoff} & \textbf{use}\\
\hline
Banana tip (either tip)                     & 22.45 & 63.27\\
Binoculars (both barrels)                & 12.50 & 93.88\\
Camera shutter button                   & 34.00 & 69.39\\
Eyeglasses (both temples)             & 4.00  & 64.58\\
Flashlight button                             & 28.00 & 62.00\\
Hammer (head)                               & 38.00 & 0.00\\
Mouse (both click buttons)            & 16.00 & 84.00\\
PS controller (both front buttons) & 2.00   & 40.81\\
PS controller (both analog sticks) & 2.00   & 22.44\\
Scissors (handle)                            & 38.00 & 100.00\\
Scissors (blade)                              & 60.00 & 0.00\\
Water-bottle cap                            & 16.00 & 67.35\\
Wine glass stem                             & 56.00 & 30.61\\
\end{tabular}
\caption{Fraction of participants that touched active areas for different functional intents. See Fig.~\ref{fig:intent_influence} for examples.}
\label{tab:intent_influence}
\end{table}

To quantify the influence of functional intent, we define `active areas' (highlighted in green in Figure~\ref{fig:intent_influence}) on the surface of some objects and show the
fraction of participants that touched that area (evidenced by the map value being greater than 0.4) in Table~\ref{tab:intent_influence}.

\begin{figure}[h!]
\centering
\includegraphics[width=\textwidth]{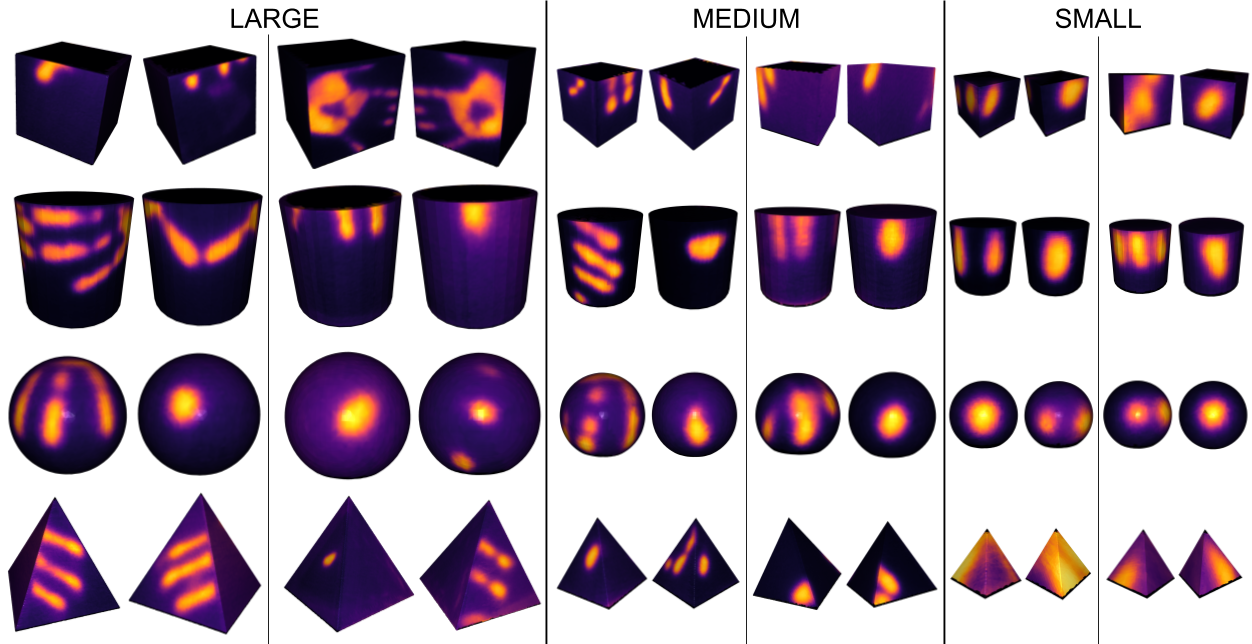}
\caption{Influence of object size on contact: Two dominant grasps for objects of same shape and varying size.}
\label{fig:size_influence}
\end{figure}

\begin{figure}
    \centering
    \includegraphics[width=0.95\textwidth]{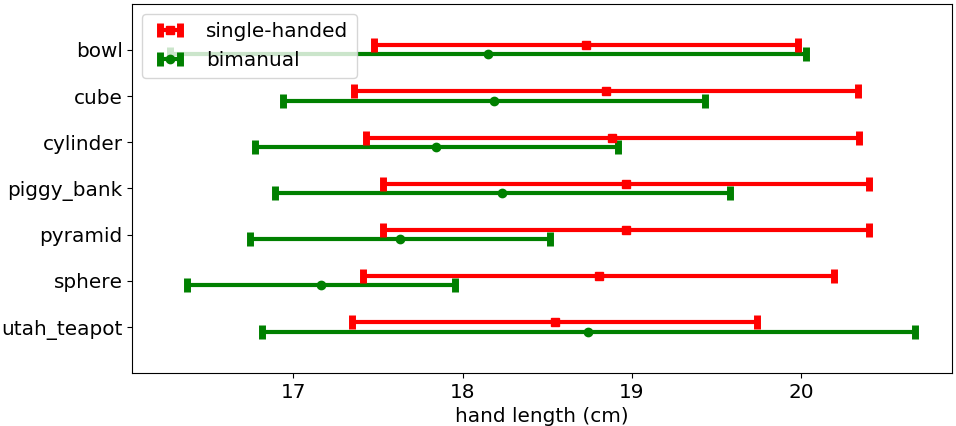}
    \caption{Relationship between hand length (wrist to mid fingertip) and single-handed/bimanual grasps. The intervals show mean and 1 standard deviation. Cube, cylinder, pyramid and sphere are of the \texttt{large} size.}
    \label{fig:bimanual_analysis}
\end{figure}

\noindent
\textbf{Effect of object size.}
Figure~\ref{fig:size_influence} shows the dominant contact maps for objects of the same shape at three different sizes. Small objects exhibit grasps with two or three fingertips, while larger objects are often grasped with more fingers and more than the fingertips in contact with the object. Grasps for large objects are bi-modal: bimanual using the full hands, or single-handed using fingertips. To quantify this, we manually labelled grasps as bimanual/single-handed, and show their relation to hand size in Fig.~\ref{fig:bimanual_analysis}. The figure shows that people with smaller hands prefer to grasp \texttt{large} objects (for `handoff') with bimanual grasps. No bimanual grasps were observed for the \texttt{medium} and \texttt{small} object sizes.

%\begin{figure*}[t!]
%\begin{subfigure}{0.05\columnwidth}
%	\includegraphics[width=\textwidth]{fingertips}
%	%\captionlistentry{}
%	\caption{}
%	\label{fig:palm_print}
%\end{subfigure}
%\begin{subfigure}{0.45\textwidth}
%	\includegraphics[width=\textwidth]{use_contact_areas}
%	\caption{Contact areas for functional intent `use'}
%	\label{fig:use_contact_areas}
%\end{subfigure}
%\begin{subfigure}{0.45\textwidth}
%	\includegraphics[width=\textwidth]{handoff_contact_areas}
%	\caption{Contact areas for functional intent `handoff'}
%	\label{fig:handoff_contact_areas}
%\end{subfigure}
%\caption{(a): Palm contact on plate, \textcolor{OliveGreen}{annotated fingertips}. (b, c): Contact areas for objects in \dsetName, averaged across participants. The \textcolor{red}{red line} indicates a loose upper bound on contact area for a fingertip-only grasp, which is doubled for objects which have bimanual grasps.}
%\label{fig:fingertip_contact}
%\end{figure*}
\begin{figure*}
    \centering
    \includegraphics[width=0.98\textwidth]{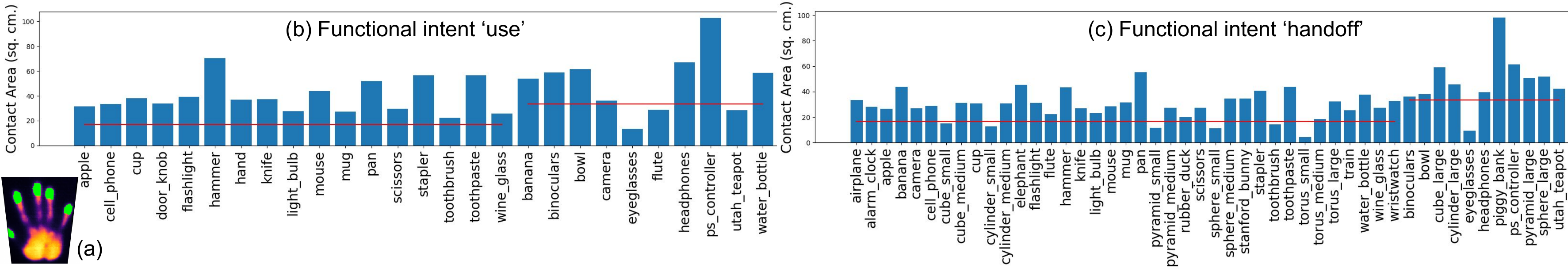}
    \caption{(a): Palm contact on plate, \textcolor{OliveGreen}{annotated fingertips}. (b, c): Contact areas for objects in \dsetName, averaged across participants. The \textcolor{red}{red line} indicates a loose upper bound on contact area for a fingertip-only grasp, which is doubled for objects which have bimanual grasps.}
    \label{fig:contact_areas}
\end{figure*}

\noindent
\textbf{How much of the contact is fingertips?}
Contact is traditionally modelled in robotics~\cite{schmidt2015depth} and simulation~\cite{grasp_from_contact} as a single point. However, the contact maps in Figures~\ref{fig:teaser}, \ref{fig:intent_influence} and \ref{fig:size_influence} show that human grasps have much more than fingertip contact. Single-point contact modeling is  inspired by the prevalence of rigid manipulators on robots, but with the recent research interest in \textit{soft robots}~\cite{soft_robots0, soft_robots1}, we now have access to manipulators that contact the object at other areas on the finger. Data in \dsetName shows the use of non-fingertip contact for highly capable soft manipulators: human hands. For each contact map, we calculated the contact area by integrating the area of all the contacted faces in the mesh. A face is contacted if any of its three vertices have a contact value greater than 0.4. Figures~\ref{fig:contact_areas}(b) and~\ref{fig:contact_areas}(c) show the contact areas for all objects under both functional intents, averaged across participants. Next, we calculated an upper bound on the contact area if only all 5 fingertips were touching the object. This was done by capturing the participants' palm print on a flat plate, where it is easy to manually annotate the fingertip regions (shown in Figure~\ref{fig:contact_areas}(a)). The total surface area of fingertips in the palm print is the desired upper bound. It was doubled for objects for which we observe bimanual grasps. This upper bound was averaged across four participants, and is shown as the red line in Figures~\ref{fig:contact_areas}(b) and~\ref{fig:contact_areas}(c). Note that this is a loose upper bound, since many real-world fingertip-only grasps don't involve all five fingertips, and we mark the entire object category as bimanual if even one participant performs a bimanual grasp. Total contact area for many objects is significantly higher than the upper bound on fingertip-only contact area, indicating the large role that the soft tissue of the human hand plays in grasping and manipulation. This motivates the inclusion of non-fingertip areas in grasp prediction and modeling algorithms, and presents an opportunity to inform the design of soft robotic manipulators. Interestingly, the average contact area for some objects (e.g. bowl, mug, PS controller, toothbrush) differs across functional intent, due to different kinds of grasps used.
\section{Predicting Contact Maps} \label{sec:prediction}
In this section, we describe experiments to predict contact maps for objects based on their shape. \dsetName is the first large scale dataset that enables training
data-intensive deep learning models for this task. Since \dsetName includes diverse contact maps for each object, the mapping from object shape to contact map is one-to-many and makes the task challenging. We explore two representations for object shape: single-view RGB-D, and full 3D. Since the contact patterns are significantly influenced by the functional intent, we train separate models for `hand-off' and `use'.

\begin{figure}
  \includegraphics[width=\textwidth]{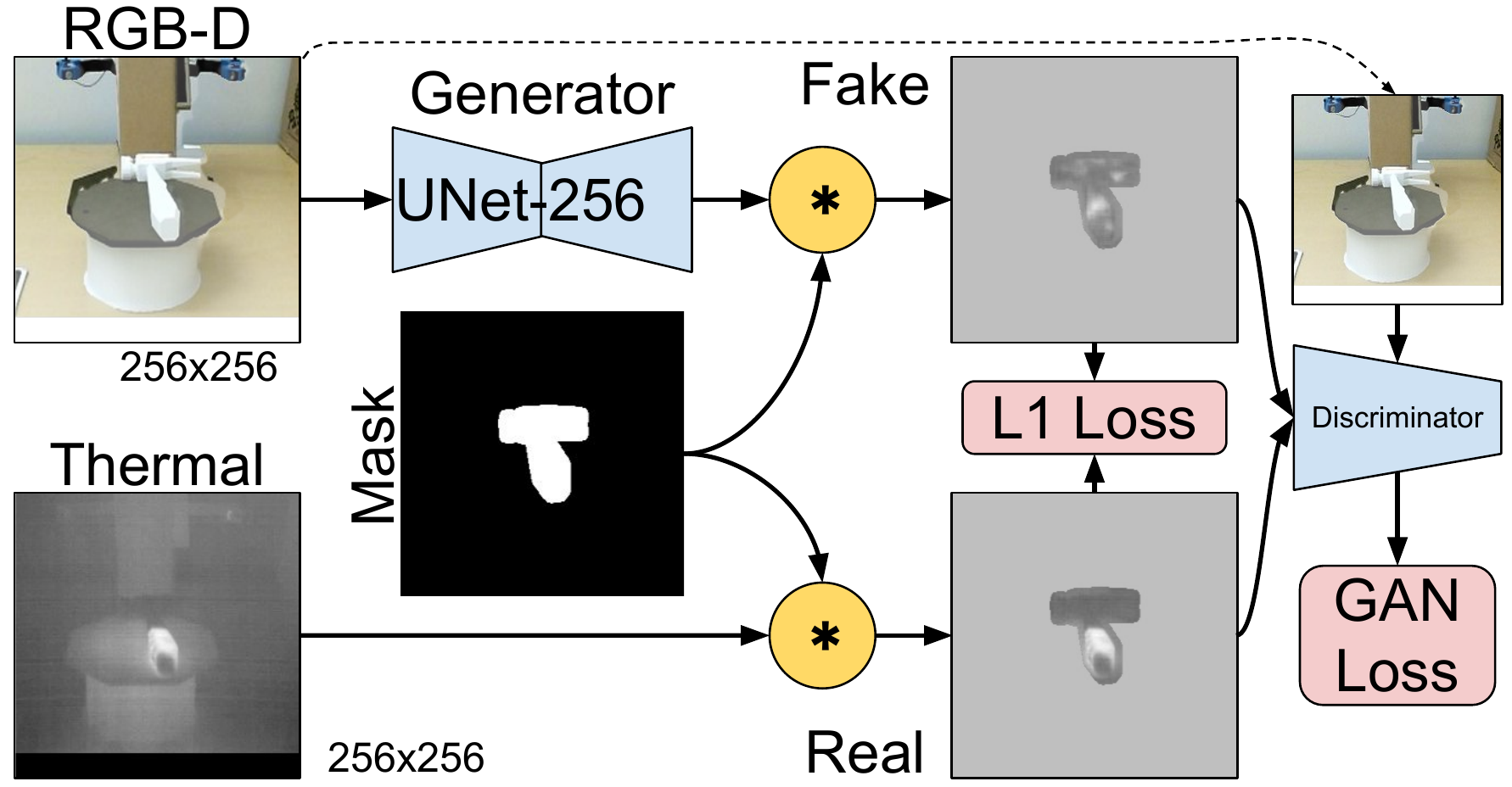}
  \caption{Training procedure for single-view contact map prediction. The discriminator has 5 conv layers followed by batch norm and leaky ReLU.}
  \label{fig:2d_pred_training}
\end{figure}

\subsection{Single-view Prediction}
Object shape is represented by an RGB-D image, and a 2D contact map is predicted for the visible part of the object. A single view
might exclude information about important aspects of the object shape, and `interesting' parts of the contact map might lie in the unseen half of the object. However, this representation has the advantage of being easily applicable to real-world robotics scenarios where mobile manipulators are often required to grasp objects after observing
them from a single view. We used generative adversarial network (GAN)-based image-to-image translation~\cite{pix2pix, cyclegan, unit} for this task, since the optimization procedure of conditional GANs is able to model a one-to-many input-output mapping~\cite{cgan, gan}.

\begin{figure}
  \includegraphics[width=\textwidth]{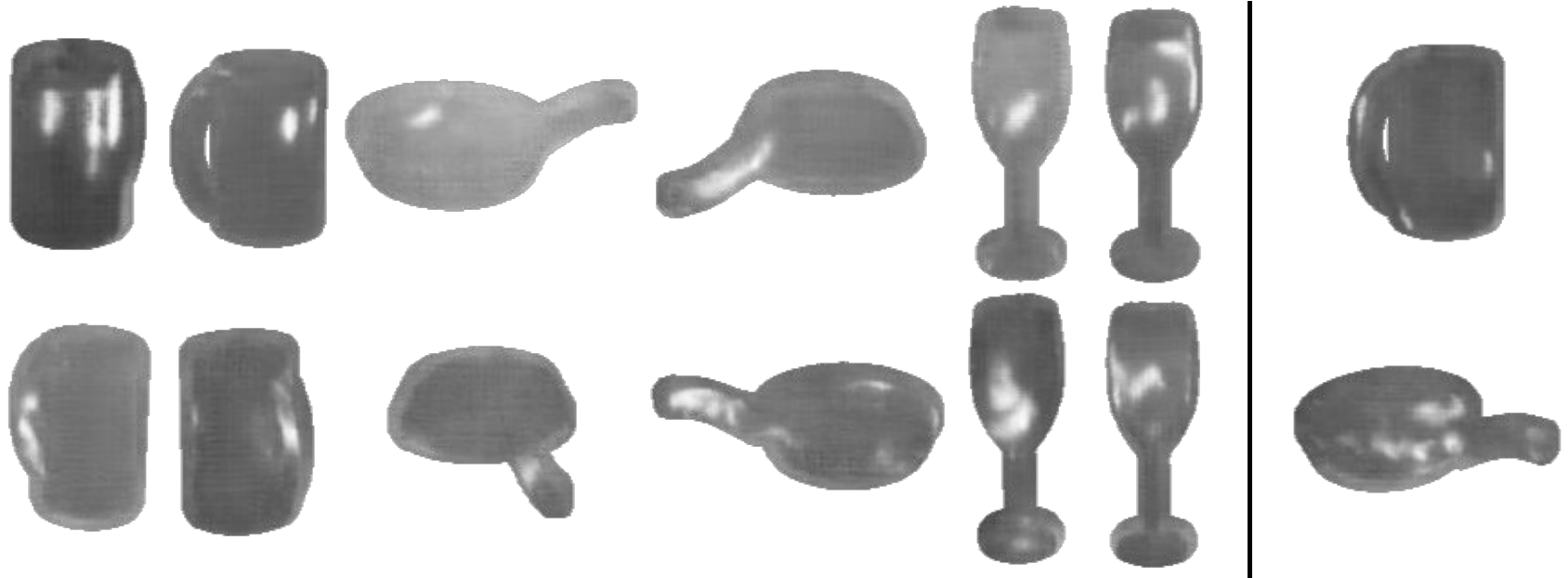}
  \caption{Single-view predictions from the pix2pix model for three \textit{unseen} object classes: mug, pan and wine glass. Top: handoff intent, bottom: use intent. Rightmost column: uninterpretable predictions.}
  \label{fig:2d_pred_results}
\end{figure}

Figure~\ref{fig:2d_pred_training} shows our training procedure and network architecture, which has roughly 54M and 3M parameters in the generator and discriminator
respectively. We modified pix2pix~\cite{pix2pix} to accept a 4-channel RGB-D input and predict a single-channel contact map. The RGB-D stream from object scanning was registered to the thermal images, and used as input. Thermal images were used as a proxy for the single-view contact map. To focus the generator and discriminator on the object, we cropped a 256$\times$320 patch around the object and masked all images by the object silhouette. All images from mug, pan, and wineglass were held out and used for testing. Figure~\ref{fig:2d_pred_results} shows some predicted contact maps for these unseen objects, selected for looking realistic. Mug predictions for use have finger contact on the handle, whereas contact is observed over the top for handoff. Pan use predictions show grasps at the handle, while handoff predictions additionally show a bi-manual grasp of the handle and side. Similarly, the wine glass indicates contact with a side grasp for use and over the opening for handoff.

\begin{figure*}[t!]
\begin{subfigure}{0.39\textwidth}
  \includegraphics[width=\textwidth]{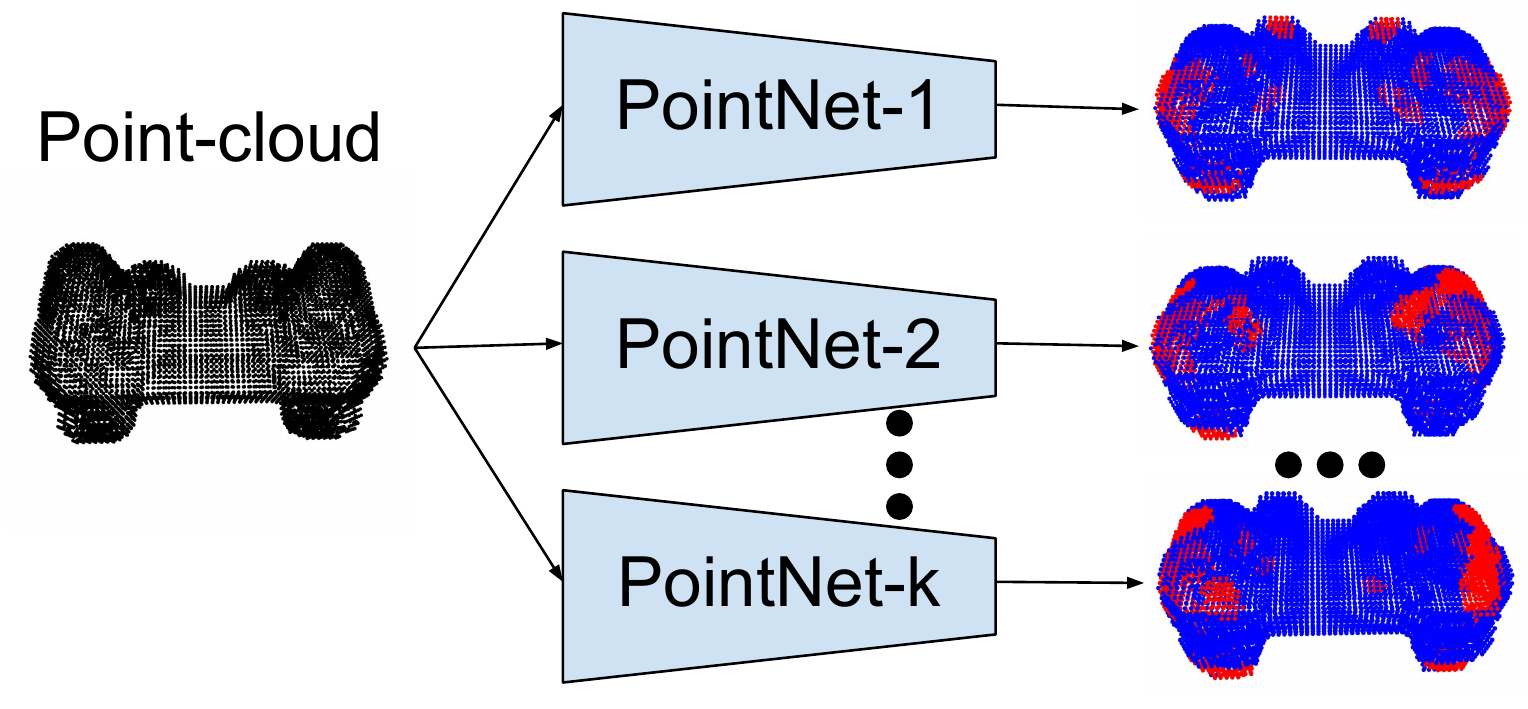}
  \caption{sMCL with a PointNet predictor}
  \label{fig:smcl_pointnet_training}
\end{subfigure} \hfill
\begin{subfigure}{0.59\textwidth}
  \includegraphics[width=\textwidth]{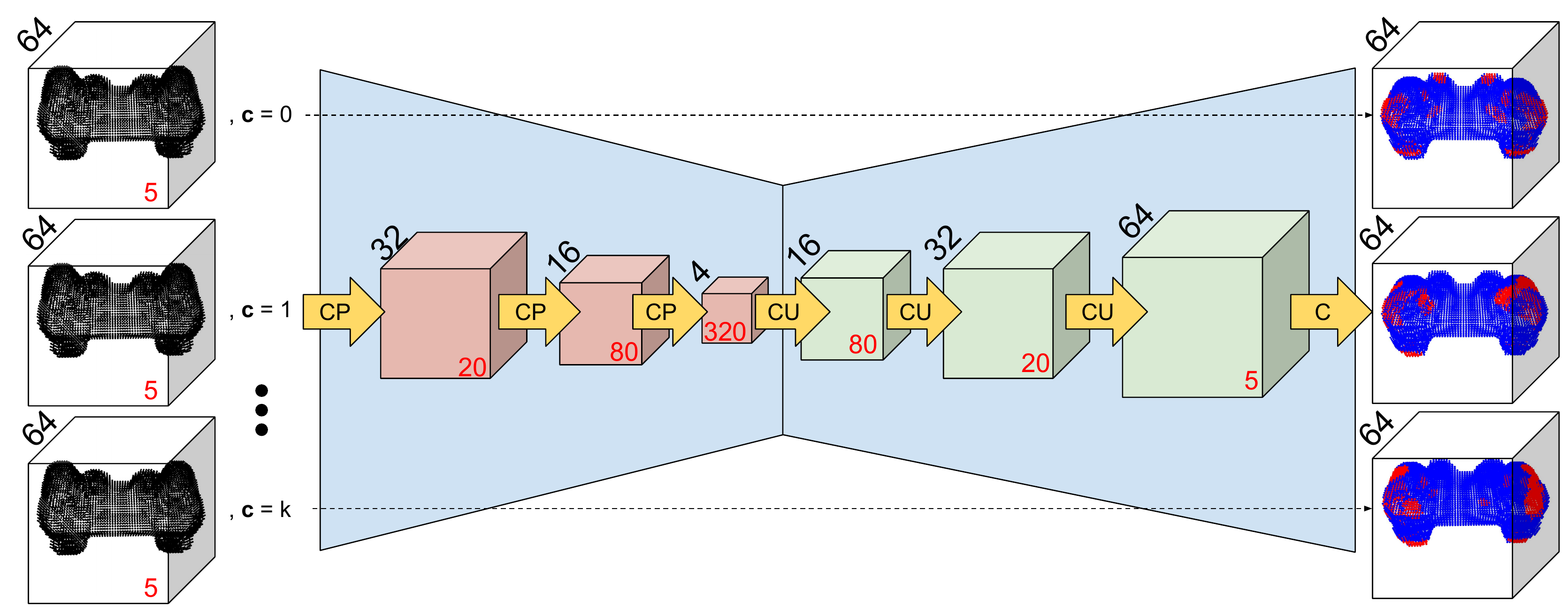}
  \caption{DiverseNet with a VoxNet predictor. CP: $3^3$ conv with batch norm, ReLU and max pooling, CU: $3^3$ conv with batch norm, ReLU
  and nearest neighbor upsampling. Black numbers: size of voxel grid, \textcolor{red}{red} numbers: number of channels.}
  \label{fig:diversenet_voxnet_training}  
\end{subfigure}
\caption{3D data representations and training strategies for predicting diverse contact maps. sMCL~\cite{smcl} requires multiple instances of a network, while
DiverseNet~\cite{diversenet} uses a single instance with an integer valued control variable. PointNet~\cite{pointnet} operates on unordered point-clouds, whereas
VoxNet~\cite{voxnet} uses voxel occupancy grids. }
\end{figure*}

\subsection{3D Prediction} \label{sec:3d_pred}
Full 3D representation gives access to the entire shape of the object, and alleviates the view-consistency problems observed during single-view prediction.

\noindent
\textbf{Learning a one-to-many-mapping}.
Stochastic Multiple Choice Learning~\cite{smcl} (sMCL) trains an ensemble of $k$ predictors to generate $k$ contact maps for each input (see Figure~\ref{fig:smcl_pointnet_training}). Each input has multiple equally correct ground truth maps. During training, the loss is backpropagated from each ground truth contact map to the network that makes the prediction closest to it. To encourage all members of the ensemble to be trained equally, as mentioned in~\cite{smcl_trick}, we made this association soft by routing the gradient to the closest network with a 0.95 weight and distributed the rest equally among other members of the ensemble, and randomly dropped entire predictions with a 0.1 probability. We trained models with $k=1$ and $k=10$.

In contrast, DiverseNet~\cite{diversenet} generates diverse predictions from a single predictor network by changing the value of a one-hot encoded control variable \textbf{c} that is concatenated to internal feature maps of the network (See Figure~\ref{fig:diversenet_voxnet_training}). Each ground truth contact map is associated with the closest prediction and gradients are routed through the  appropriate \textbf{c} value. Diverse predictions can be generated at test time by varying \textbf{c}. Compared to sMCL, DiverseNet requires significantly fewer trainable parameters. We used 10 one-hot encoded \textbf{c} values in our experiments.

\noindent
\textbf{3D representation}. We represented the 3D object shape in two forms: pointcloud and voxel occupancy grid. PointNet~\cite{pointnet} operates on a pointcloud representation of the object shape, with points randomly sampled from the object surface. We normalized the XYZ position of each point to fit the object in a unit cube. The XYZ position and the normalization scale factor were used as 4-element features for each point. The network was trained by cross entropy loss to predict whether each voxel is in contact. We used a PointNet architecture with a single T-Net and 1.2M parameters.

VoxNet~\cite{voxnet} operates on a solid occupancy grid of the object in a $64^3$ voxelized space, and predicts whether each voxel is contacted. It uses 3D convolutions to learn shape features. The four features used for PointNet were used in addition to the binary occupancy value to form a 5-element feature vector for each voxel. Cross entropy loss was enforced only on the voxels on the object surface. The network architecture is shown in Figure~\ref{fig:diversenet_voxnet_training}, and has approximately 1.2M parameters.

\begin{table*}[h!]
\centering
\resizebox{0.99\textwidth}{!}{
%\small
\begin{tabular}{c|c|c|c|c|c|c||c|c|c|c|c|c}
\multirow{3}{*}{Test object} & \multicolumn{6}{c||}{Handoff} & \multicolumn{6}{c}{Use}\\
\cline{2-13}
& \multicolumn{2}{|c|}{sMCL ($k=1$)} & \multicolumn{2}{|c|}{sMCL ($k=10$)} & \multicolumn{2}{|c||}{DiverseNet ($k=10$)}
& \multicolumn{2}{|c|}{sMCL ($k=1$)} & \multicolumn{2}{|c|}{sMCL ($k=10$)} & \multicolumn{2}{|c}{DiverseNet ($k=10$)}\\
\cline{2-13}
& VoxNet & PointNet & VoxNet & PointNet & VoxNet & PointNet & VoxNet & PointNet & VoxNet & PointNet & VoxNet & PointNet\\
\hline
pan            & 76.80  & - & \textbf{7.13}   & 20.43 & 8.48                & 19.68 & 17.22  & - & 8.25   & 43.57 & \textbf{5.12}  & 22.58\\
wine glass & 59.37  & - & \textbf{11.11}  & 14.59 & 28.69               & 17.28 & 50.18  & - & 11.06 & 14.79  & 13.98             & \textbf{10.47}\\
mug           & 29.93  & - & 16.68              & 27.10 & \textbf{15.77}  & 21.60 & 66.03  & - & 32.51 & 31.30 & \textbf{7.06} & 32.41\\ \hline
average     & 55.37  & - & \textbf{11.64} & 20.71 & 17.65                & 19.52 & 44.48 & - & 17.27  & 29.89 & \textbf{8.72} & 21.82\\
\hline
\end{tabular}}
\caption{Diverse 3D contact map prediction errors (\%) for the models presented in Section~\ref{sec:3d_pred}. Errors were calculated by matching each ground truth contact map
with the closest from $k$ diverse predictions, discarding predictions with no contact. `-' indicates that no contact was predicted.}
\label{tab:3d_pred_results}
\end{table*}

\begin{figure*}[h!]
\begin{subfigure}{0.45\textwidth}
  \includegraphics[width=\textwidth]{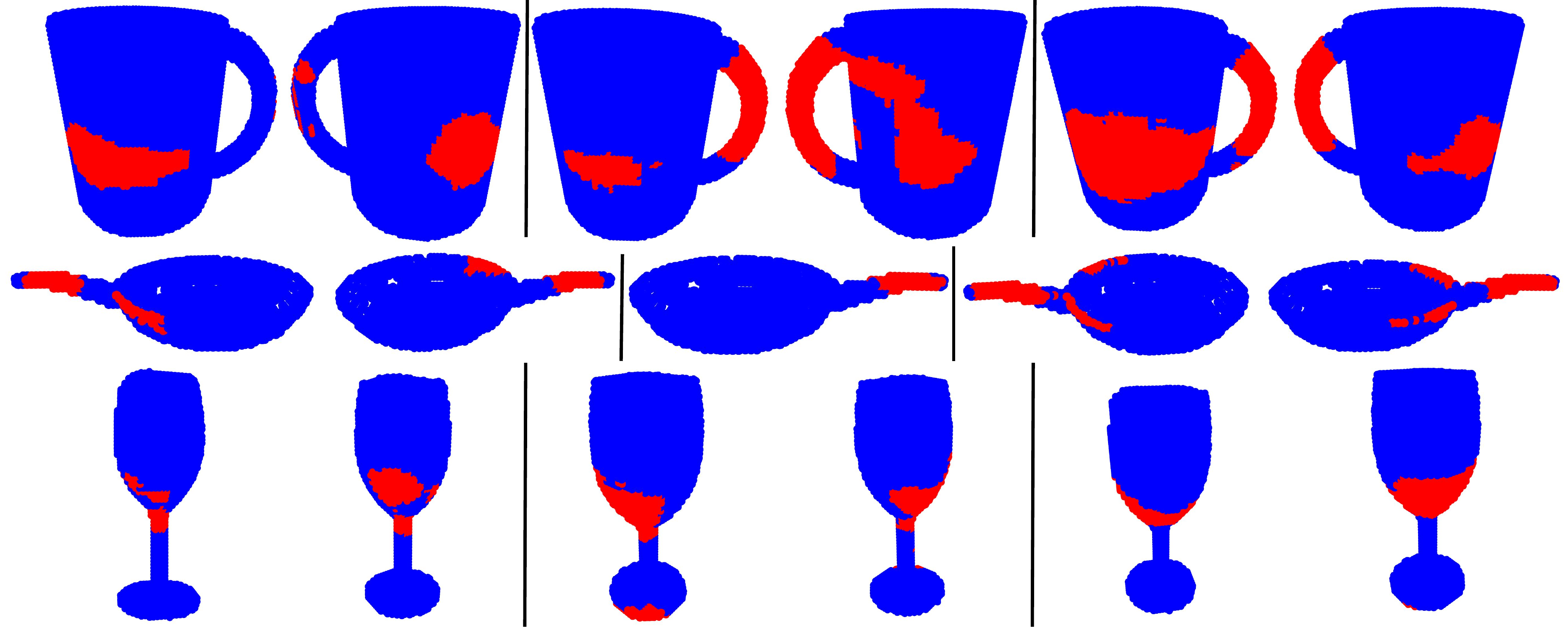}
  \caption{Contact map predictions for unseen object classes}
  \label{fig:3d_pred_results}
\end{subfigure}
\begin{subfigure}{0.53\textwidth}
  \includegraphics[width=\textwidth]{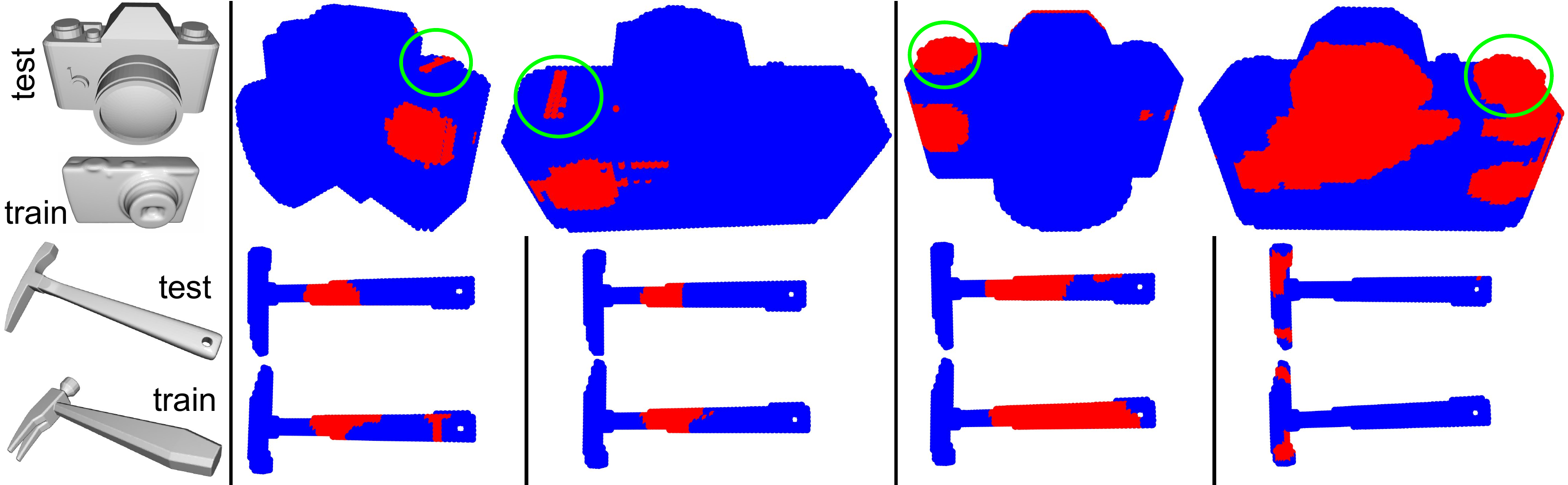}
  \caption{Contact map predictions for an unseen shape of training object classes}
  \label{fig:3d_pred_results_newobjects}  
\end{subfigure}
\caption{Two views of diverse 3D contact map predictions. (a) \textit{Unseen} object classes: mug, pan, and wine glass, (b) \textit{Unseen shape} of training object classes: camera and hammer. Intent: use, Model: VoxNet-DiverseNet, \textcolor{red}{Red}: contact.}
\end{figure*}

\noindent
\textbf{Experiments}
We conducted experiments with both VoxNet and PointNet, using the sMCL and DiverseNet strategies for learning a one-to-many-mapping. For DiverseNet, we concatenated $\mathbf{c}$ to the output of the first and fifth conv layers in VoxNet, and to the input transformed by T-Net and the output of the second-last MLP
in PointNet. Voxelization of the meshes was done using the algorithm of~\cite{binvox_algo} implemented in binvox~\cite{binvox_website}. The PointNet input was generated by randomly sampling 3000 points from the object surface. We thresholded the contact maps at 0.4 after applying the sigmoid described in Section~\ref{sec:analysis}, to generate ground truth for classification. We augmented the dataset by randomly rotating the object around the yaw axis. PointNet input was also augmented by randomly choosing an axis and scaling the points along that axis by a random factor in  [0.6, 1.4]. Dropout with $p=0.2$ was applied to VoxNet-DiverseNet input. We found that similar dropout did not improve results for other models. Random sampling of surface points automatically acts like dropout for PointNet models, and sMCL models already incorporate a different dropout strategy as mentioned in Section~\ref{sec:3d_pred}.  The cross entropy loss for contacted voxels was weighted by a factor of 10, to account for class imbalance. All models were trained with SGD with a learning rate of 0.1, momentum of 0.9 and weight decay of 5e-4. Batch size was 5 for models with $k=10$, and 25 for models with $k=1$.

Table~\ref{tab:3d_pred_results} shows results on held-out test objects (mug, pan and wine glass). We conclude that the voxel occupancy grid representation is better for this task, and that a model limited to making a single prediction does not capture the complexity in \dsetName. Figures~\ref{fig:3d_pred_results} and~\ref{fig:3d_pred_results_newobjects} show some of the `use' intent predictions for unseen object classes and unseen shapes of training object classes respectively, selected for looking realistic. Mug predictions show horizontal grasps around the body. Predictions for the pan are concentrated at the handle, with one grasp being bimanual. Wine glass predictions show grasps at the body-stem intersection. Camera predictions show contact at the shutter button and sides, while predictions for the hammer show contact at the handle (and once at the head).
\section{Conclusion and Future Work} \label{sec:conclusion}
We presented \dsetName, the first large-scale dataset of contact maps from functional grasping, analyzed the data to reveal interesting aspects of grasping behavior, and explored data representations and training strategies for predicting contact maps from object shape. We hope to spur future work in multiple areas. Contact patterns could inform the design of soft robotic manipulators by aiming to be able to cover object regions touched by humans. Research indicates that in some situations hand pose can be guided by contact points~\cite{grasp_from_contact, sridhar2016real}. Using contact maps to recover and/or assist in predicting the hand pose in functional grasping is an exciting problem for future research.

\textbf{Acknowledgements}: We thank Varun Agrawal for lending the 3D printer, Ari Kapusta for initial discussions on thermal cameras, NVIDIA for a GPU grant, and all the anonymous participants involved in data collection.
\section*{Supplementary Material}

%%%%%%%%% ABSTRACT
\subsection*{Abstract}
This document provides supplementary material for our submission. We compare ContactDB heatmaps qualitatively against the crowdsourced tactile saliency maps from~\cite{tactile_mesh_saliency}. We discuss the extent of heat dissipation while scanning the object, and potential sources of error in observing contact through the thermal camera and the texture mapping process. Lastly, we list the 50 objects used in \dsetName and the instructions given to participants for grasping the subset of 27 objects with the `use' post-grasp intent. \dsetName can be explored interactively at \url{https://contactdb.cc.gatech.edu}.

\subsection*{Comparison to Tactile Mesh Saliency~\cite{tactile_mesh_saliency}}
\begin{figure}[h!]
    \centering
    \includegraphics[width=0.98\textwidth]{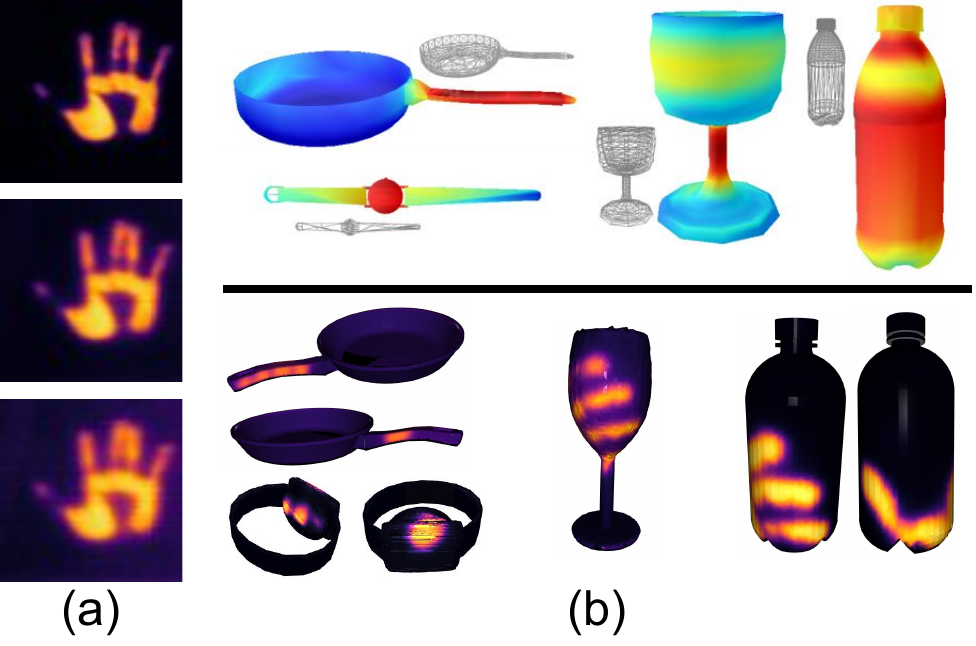}
    \caption{(a) Heat dissipation in the thermal images. Top-bottom: 0s, 18s, 35s. (b) Contact information collected by online crowd-sourcing (\cite{tactile_mesh_saliency}, top row) and ContactDB (ours, bottom row).}
    \label{fig:main}
\end{figure}

Qualitatively, the closest work to ContactDB that we've found is~\cite{tactile_mesh_saliency}, which collects contact saliency information through crowd-sourcing by pairwise comparison of surface points. Figure~\ref{fig:main}(b) compares common objects from both datasets. Notably, data from~\cite{tactile_mesh_saliency} lacks clear finger-marks and resembles averaged contact maps. That data may be less accurate because it relies on self-reporting. For example, our data shows that people rarely contact the bottom half of the wine glass stem, whereas~\cite{tactile_mesh_saliency} shows high saliency for the entire stem.

\subsection*{Heat Dissipation During Data Collection}
Scanning takes 18 s for a \ang{360} rotation. Owing to the consistent use of hand-warmers and PLA material for 3D printed objects, thermal prints take more than 35 s to diffuse significantly (See Fig.~\ref{fig:main}(a)). Heat conduction across the surface of the plate does not seem to be a significant source of variation between 0 s and 18 s, since the prints are comparable in size and lack strongly blurred edges. This shows that the dissipation of finger heat on the object surface produces minimal artifacts in the contact maps presented in the paper. We operate the turntable motor at the maximum possible speed that avoids high centrifugal force and wear-and-tear.

\subsection*{Accuracy of Texture Mapping}
\begin{figure}
\includegraphics[width=0.5\textwidth]{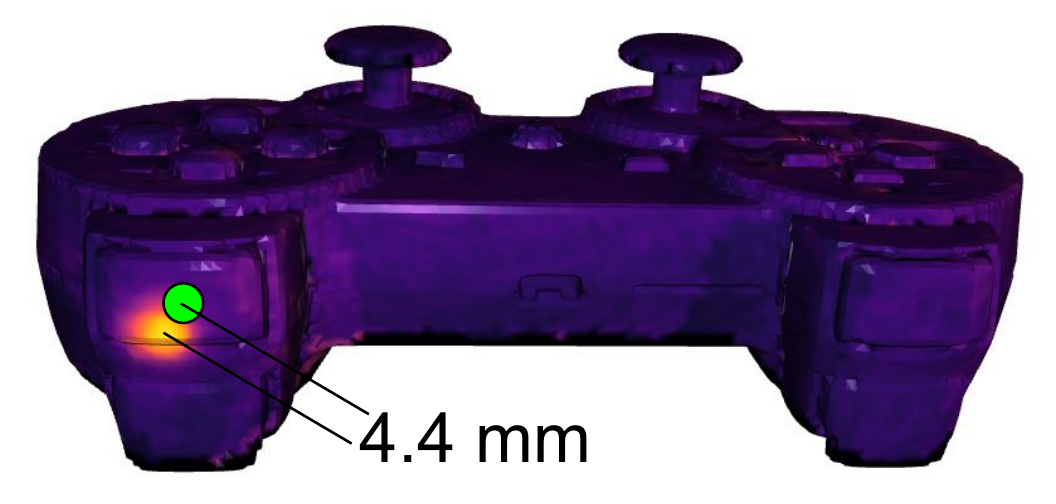}
\caption{Geometric error of the texture mapping process. The spot on the front button shown in \textcolor{OliveGreen}{green} was precision-heated with a warm pencil-top eraser.}
\label{fig:geometric_error}
\end{figure}

As discussed in Section 3.3 of the paper, thermal images from 9 views and corresponding object pose estimates are used in a texture mapping algorithm to produce a final mesh 
textured with a contact map. The whole process has multiple potential sources of error: calibration of the intrinsics and extrinsics of the Kinect v2 and thermal camera,
inaccuracy in 3D printing the object, errors in object pose estimates due to noise/distortion in the Kinect depth maps, artifacts introduced by the texture mapping algorithm, etc.
As such, the accuracy of this process can be different for different objects and sessions. In Figure~\ref{fig:geometric_error}, we attempt to quantify this error for one instance
where we precisely heated a spot on the front button of the PS controller using a heated pencil-top eraser. In this case, we observed a final geometric error of 4.4 mm.

\subsection*{List of Objects}
Table~\ref{tab:object_list} shows a list of all 50 objects in \dsetName, along with information about the which of these objects are included in the two functional grasping 
categories, and the specific `use' instructions.

\begin{table*}
\centering
\begin{tabular}{c|c|c|c}
\textbf{Object} & \textbf{handoff} & \textbf{use} & \textbf{use instruction}\\
\hline
airplane & \checkmark & &\\
alarm clock & \checkmark	 & &\\
apple & \checkmark & \checkmark & eat\\
banana	& \checkmark & \checkmark & peel\\
binoculars & \checkmark & \checkmark & see through\\
bowl & \checkmark & \checkmark & drink from\\
camera & \checkmark & \checkmark & take picture\\
cell phone & \checkmark & \checkmark & talk on\\
cube (small)  & \checkmark & &\\
cube (medium) & \checkmark & &\\	
cube (large) & \checkmark & &\\
cup & \checkmark & \checkmark & drink from\\
cylinder (small) & \checkmark &&\\
cylinder (medium)	&\checkmark &&\\
cylinder (large) & \checkmark & &\\
door knob	& & \checkmark & twist to open door\\
elephant & \checkmark	&&\\
eyeglasses & \checkmark & \checkmark & wear\\
flashlight & \checkmark & \checkmark & turn on\\
flute & \checkmark & \checkmark & play\\
hammer & \checkmark & \checkmark & hit a nail\\
hand & & \checkmark & shake\\
headphones & \checkmark & \checkmark & wear\\
knife & \checkmark & \checkmark & cut\\
light bulb & \checkmark & \checkmark & screw in a socket\\
mouse & \checkmark	 & \checkmark & use to point and click\\
mug & \checkmark & \checkmark & drink from\\
pan & \checkmark & \checkmark & cook in\\
piggy bank & \checkmark &&\\
PS controller & \checkmark & \checkmark & play a game with\\
pyramid (small)	& \checkmark	&&\\
pyramid (medium)	& \checkmark	&&\\
pyramid (large)	& \checkmark	&& \\
rubber duck & \checkmark &&\\
scissors & \checkmark & \checkmark & cut with\\
sphere (small) & \checkmark	&&\\
sphere (medium) & \checkmark	&&\\
sphere (large) & \checkmark	&&\\
Stanford bunny & \checkmark & &\\
stapler & \checkmark & \checkmark & staple\\
toothbrush & \checkmark & \checkmark & brush teeth\\
toothpaste & \checkmark &\checkmark & squeeze out toothpaste\\
torus (small) & \checkmark	&&\\
torus (medium) &\checkmark	&&\\
torus (large) & \checkmark	&&\\
train & \checkmark &&\\
Utah teapot & \checkmark & \checkmark & pour tea from\\
water bottle & \checkmark & \checkmark & open\\
wine glass	 & \checkmark & \checkmark & drink wine from\\
wristwatch & \checkmark & &\\
\hline
\textbf{Total} & \textbf{48} & \textbf{27} &\\
\end{tabular}
\caption{List of objects in \dsetName and specific `use' instructions}
\label{tab:object_list}
\end{table*}

\FloatBarrier

\Urlmuskip=0mu plus 1mu\relax
{\small
\bibliographystyle{ieee}
\bibliography{references}

\begin{thebibliography}{10}\itemsep=-1pt

\bibitem{tactile_logging}
Shuichi Akizuki and Yoshimitsu Aoki.
\newblock Tactile logging for understanding plausible tool use based on human
  demonstration.
\newblock In {\em British Machine Vision Conference 2018, {BMVC} 2018,
  Northumbria University, Newcastle, UK, September 3-6, 2018}, page 334, 2018.

\bibitem{intent_influence0}
Caterina Ansuini, Livia Giosa, Luca Turella, Gianmarco Alto{\`e}, and Umberto
  Castiello.
\newblock An object for an action, the same object for other actions: effects
  on hand shaping.
\newblock {\em Experimental Brain Research}, 185(1):111--119, 2008.

\bibitem{manual_ann_grasp1}
Ravi Balasubramanian, Ling Xu, Peter~D Brook, Joshua~R Smith, and Yoky
  Matsuoka.
\newblock Physical human interactive guidance: Identifying grasping principles
  from human-planned grasps.
\newblock {\em IEEE Transactions on Robotics}, 4(28):899--910, 2012.

\bibitem{glove_tactile_sensors}
Keni Bernardin, Koichi Ogawara, Katsushi Ikeuchi, and Ruediger Dillmann.
\newblock A sensor fusion approach for recognizing continuous human grasping
  sequences using hidden markov models.
\newblock {\em IEEE Transactions on Robotics}, 21(1):47--57, 2005.

\bibitem{icp}
PJ Besl and Neil~D McKay.
\newblock A method for registration of 3-d shapes.
\newblock {\em Pattern Analysis and Machine Intelligence, IEEE Transactions
  on}, 14(2):239--256, 1992.

\bibitem{tax_ann_grasp2}
Ian~M Bullock, Thomas Feix, and Aaron~M Dollar.
\newblock The yale human grasping dataset: Grasp, object, and task data in
  household and machine shop environments.
\newblock {\em The International Journal of Robotics Research}, 34(3):251--255,
  2015.

\bibitem{ycb}
Berk Calli, Aaron Walsman, Arjun Singh, Siddhartha Srinivasa, Pieter Abbeel,
  and Aaron~M Dollar.
\newblock Benchmarking in manipulation research: The ycb object and model set
  and benchmarking protocols.
\newblock {\em arXiv preprint arXiv:1502.03143}, 2015.

\bibitem{grasp_influence}
Umberto Castiello.
\newblock The neuroscience of grasping.
\newblock {\em Nature Reviews Neuroscience}, 6(9):726, 2005.

\bibitem{choi_grasping}
Changhyun Choi, Wilko Schwarting, Joseph DelPreto, and Daniela Rus.
\newblock Learning object grasping for soft robot hands.
\newblock {\em IEEE Robotics and Automation Letters}, 2018.

\bibitem{amazon_picking_challenge}
Nikolaus Correll, Kostas~E Bekris, Dmitry Berenson, Oliver Brock, Albert Causo,
  Kris Hauser, Kei Okada, Alberto Rodriguez, Joseph~M Romano, and Peter~R
  Wurman.
\newblock Analysis and observations from the first amazon picking challenge.
\newblock {\em IEEE Transactions on Automation Science and Engineering},
  15(1):172--188, 2018.

\bibitem{size_influence}
Raymond~H Cuijpers, Jeroen~BJ Smeets, and Eli Brenner.
\newblock On the relation between object shape and grasping kinematics.
\newblock {\em Journal of Neurophysiology}, 91(6):2598--2606, 2004.

\bibitem{grasp_taxonomy0}
Mark~R Cutkosky.
\newblock On grasp choice, grasp models, and the design of hands for
  manufacturing tasks.
\newblock {\em IEEE Transactions on robotics and automation}, 5(3):269--279,
  1989.

\bibitem{soft_robots0}
Raphael Deimel and Oliver Brock.
\newblock A novel type of compliant and underactuated robotic hand for
  dexterous grasping.
\newblock {\em The International Journal of Robotics Research},
  35(1-3):161--185, 2016.

\bibitem{diversenet}
Michael Firman, Neill~DF Campbell, Lourdes Agapito, and Gabriel~J Brostow.
\newblock Diversenet: When one right answer is not enough.
\newblock In {\em Proceedings of the IEEE Conference on Computer Vision and
  Pattern Recognition}, pages 5598--5607, 2018.

\bibitem{soft_robots1}
Kevin~C Galloway, Kaitlyn~P Becker, Brennan Phillips, Jordan Kirby, Stephen
  Licht, Dan Tchernov, Robert~J Wood, and David~F Gruber.
\newblock Soft robotic grippers for biological sampling on deep reefs.
\newblock {\em Soft robotics}, 3(1):23--33, 2016.

\bibitem{garcia2018first}
Guillermo Garcia-Hernando, Shanxin Yuan, Seungryul Baek, and Tae-Kyun Kim.
\newblock First-person hand action benchmark with rgb-d videos and 3d hand pose
  annotations.
\newblock In {\em Proceedings of the IEEE Conference on Computer Vision and
  Pattern Recognition}, pages 409--419, 2018.

\bibitem{ghazaei2018dealing}
Ghazal Ghazaei, Iro Laina, Christian Rupprecht, Federico Tombari, Nassir Navab,
  and Kianoush Nazarpour.
\newblock Dealing with ambiguity in robotic grasping via multiple predictions.
\newblock {\em arXiv preprint arXiv:1811.00793}, 2018.

\bibitem{gan}
Ian Goodfellow, Jean Pouget-Abadie, Mehdi Mirza, Bing Xu, David Warde-Farley,
  Sherjil Ozair, Aaron Courville, and Yoshua Bengio.
\newblock Generative adversarial nets.
\newblock In {\em Advances in neural information processing systems}, pages
  2672--2680, 2014.

\bibitem{hamer2010object}
Henning Hamer, Juergen Gall, Thibaut Weise, and Luc Van~Gool.
\newblock An object-dependent hand pose prior from sparse training data.
\newblock In {\em 2010 IEEE Computer Society Conference on Computer Vision and
  Pattern Recognition}, pages 671--678. IEEE, 2010.

\bibitem{glove_ann_grasp0}
Guido Heumer, Heni~Ben Amor, Matthias Weber, and Bernhard Jung.
\newblock Grasp recognition with uncalibrated data gloves-a comparison of
  classification methods.
\newblock In {\em Virtual Reality Conference, 2007. VR'07. IEEE}, pages 19--26.
  IEEE, 2007.

\bibitem{vision_grasp_tax_pred2}
De-An Huang, Minghuang Ma, Wei-Chiu Ma, and Kris~M. Kitani.
\newblock How do we use our hands? discovering a diverse set of common grasps.
\newblock In {\em The IEEE Conference on Computer Vision and Pattern
  Recognition (CVPR)}, June 2015.

\bibitem{pix2pix}
Phillip Isola, Jun-Yan Zhu, Tinghui Zhou, and Alexei~A Efros.
\newblock Image-to-image translation with conditional adversarial networks.
\newblock In {\em 2017 IEEE Conference on Computer Vision and Pattern
  Recognition (CVPR)}, pages 5967--5976. IEEE, 2017.

\bibitem{grasp_taxonomy1}
Noriko Kamakura, Michiko Matsuo, Harumi Ishii, Fumiko Mitsuboshi, and Yoriko
  Miura.
\newblock Patterns of static prehension in normal hands.
\newblock {\em American Journal of Occupational Therapy}, 34(7):437--445, 1980.

\bibitem{k_medoids}
Leonard Kaufman and Peter Rousseeuw.
\newblock {\em Clustering by means of medoids}.
\newblock North-Holland, 1987.

\bibitem{Larson:2011:HTI:1978942.1979317}
Eric Larson, Gabe Cohn, Sidhant Gupta, Xiaofeng Ren, Beverly Harrison, Dieter
  Fox, and Shwetak Patel.
\newblock Heatwave: Thermal imaging for surface user interaction.
\newblock In {\em Proceedings of the SIGCHI Conference on Human Factors in
  Computing Systems}, CHI '11, pages 2565--2574, New York, NY, USA, 2011. ACM.

\bibitem{tactile_mesh_saliency}
Manfred Lau, Kapil Dev, Weiqi Shi, Julie Dorsey, and Holly Rushmeier.
\newblock Tactile mesh saliency.
\newblock {\em ACM Transactions on Graphics (TOG)}, 35(4):52, 2016.

\bibitem{smcl}
Stefan Lee, Senthil Purushwalkam~Shiva Prakash, Michael Cogswell, Viresh
  Ranjan, David Crandall, and Dhruv Batra.
\newblock Stochastic multiple choice learning for training diverse deep
  ensembles.
\newblock In {\em Advances in Neural Information Processing Systems}, pages
  2119--2127, 2016.

\bibitem{saxena_grasping}
Ian Lenz, Honglak Lee, and Ashutosh Saxena.
\newblock Deep learning for detecting robotic grasps.
\newblock {\em The International Journal of Robotics Research},
  34(4-5):705--724, 2015.

\bibitem{glove_ann_grasp1}
Yun Lin and Yu Sun.
\newblock Grasp planning based on strategy extracted from demonstration.
\newblock In {\em Intelligent Robots and Systems (IROS 2014), 2014 IEEE/RSJ
  International Conference on}, pages 4458--4463. IEEE, 2014.

\bibitem{unit}
Ming-Yu Liu, Thomas Breuel, and Jan Kautz.
\newblock Unsupervised image-to-image translation networks.
\newblock In {\em Advances in Neural Information Processing Systems}, pages
  700--708, 2017.

\bibitem{luo2017scene}
Rachel Luo, Ozan Sener, and Silvio Savarese.
\newblock Scene semantic reconstruction from egocentric rgb-d-thermal videos.
\newblock In {\em 2017 International Conference on 3D Vision (3DV)}, pages
  593--602. IEEE, 2017.

\bibitem{dexnet}
Jeffrey Mahler, Jacky Liang, Sherdil Niyaz, Michael Laskey, Richard Doan, Xinyu
  Liu, Juan~Aparicio Ojea, and Ken Goldberg.
\newblock Dex-net 2.0: Deep learning to plan robust grasps with synthetic point
  clouds and analytic grasp metrics.
\newblock {\em arXiv preprint arXiv:1703.09312}, 2017.

\bibitem{voxnet}
Daniel Maturana and Sebastian Scherer.
\newblock Voxnet: A 3d convolutional neural network for real-time object
  recognition.
\newblock In {\em Intelligent Robots and Systems (IROS), 2015 IEEE/RSJ
  International Conference on}, pages 922--928. IEEE, 2015.

\bibitem{binvox_website}
Patrick Min.
\newblock binvox.
\newblock \url{http://www.patrickmin.com/binvox}, 2004 - 2017.
\newblock Accessed: 2018-11-16.

\bibitem{cgan}
Mehdi Mirza and Simon Osindero.
\newblock Conditional generative adversarial nets.
\newblock {\em CoRR}, abs/1411.1784, 2014.

\bibitem{tax_ann_grasp0}
Yuzuko~C Nakamura, Daniel~M Troniak, Alberto Rodriguez, Matthew~T Mason, and
  Nancy~S Pollard.
\newblock The complexities of grasping in the wild.
\newblock In {\em Humanoid Robotics (Humanoids), 2017 IEEE-RAS 17th
  International Conference on}, pages 233--240. IEEE, 2017.

\bibitem{binvox_algo}
Fakir~S. Nooruddin and Greg Turk.
\newblock Simplification and repair of polygonal models using volumetric
  techniques.
\newblock {\em IEEE Transactions on Visualization and Computer Graphics},
  9(2):191--205, 2003.

\bibitem{pham2018hand}
Tu-Hoa Pham, Nikolaos Kyriazis, Antonis~A Argyros, and Abderrahmane Kheddar.
\newblock Hand-object contact force estimation from markerless visual tracking.
\newblock {\em IEEE transactions on pattern analysis and machine intelligence},
  40(12):2883--2896, 2018.

\bibitem{puhlmann2016compact}
Steffen Puhlmann, Fabian Heinemann, Oliver Brock, and Marianne Maertens.
\newblock A compact representation of human single-object grasping.
\newblock In {\em 2016 IEEE International Conference on Intelligent Robots and
  Systems (IROS)}, page 1954–1959. IEEE, 2016.

\bibitem{pointnet}
Charles~R Qi, Hao Su, Kaichun Mo, and Leonidas~J Guibas.
\newblock Pointnet: Deep learning on point sets for 3d classification and
  segmentation.
\newblock In {\em Proceedings of the IEEE Conference on Computer Vision and
  Pattern Recognition}, pages 652--660, 2017.

\bibitem{ros}
Morgan Quigley, Josh Faust, Tully Foote, and Jeremy Leibs.
\newblock Ros: an open-source robot operating system.

\bibitem{rogez2015understanding}
Gr{\'e}gory Rogez, James~S Supancic, and Deva Ramanan.
\newblock Understanding everyday hands in action from rgb-d images.
\newblock In {\em 2015 IEEE International Conference on Computer Vision
  (ICCV)}, pages 3889--3897. IEEE, 2015.

\bibitem{smcl_trick}
Christian Rupprecht, Iro Laina, Robert DiPietro, Maximilian Baust, Federico
  Tombari, Nassir Navab, and Gregory~D Hager.
\newblock Learning in an uncertain world: Representing ambiguity through
  multiple hypotheses.
\newblock In {\em Proceedings of the IEEE International Conference on Computer
  Vision}, pages 3591--3600, 2017.

\bibitem{pcl}
Radu~Bogdan Rusu and Steve Cousins.
\newblock {3D is here: Point Cloud Library (PCL)}.
\newblock In {\em {IEEE International Conference on Robotics and Automation
  (ICRA)}}, Shanghai, China, May 9-13 2011.

\bibitem{intent_influence1}
Luisa Sartori, Elisa Straulino, and Umberto Castiello.
\newblock How objects are grasped: the interplay between affordances and
  end-goals.
\newblock {\em PloS one}, 6(9):e25203, 2011.

\bibitem{tax_ann_grasp1}
Artur Saudabayev, Zhanibek Rysbek, Raykhan Khassenova, and Huseyin~Atakan
  Varol.
\newblock Human grasping database for activities of daily living with depth,
  color and kinematic data streams.
\newblock {\em Scientific data}, 5, 2018.

\bibitem{schmidt2015depth}
Tanner Schmidt, Katharina Hertkorn, Richard Newcombe, Zoltan Marton, Michael
  Suppa, and Dieter Fox.
\newblock Depth-based tracking with physical constraints for robot
  manipulation.
\newblock In {\em 2015 IEEE International Conference on Robotics and Automation
  (ICRA)}, pages 119--126. IEEE, 2015.

\bibitem{sridhar2016real}
Srinath Sridhar, Franziska Mueller, Michael Zollh{\"o}fer, Dan Casas, Antti
  Oulasvirta, and Christian Theobalt.
\newblock Real-time joint tracking of a hand manipulating an object from rgb-d
  input.
\newblock In {\em European Conference on Computer Vision}, pages 294--310.
  Springer, 2016.

\bibitem{manual_ann_grasp0}
Danhang Tang, Hyung Jin~Chang, Alykhan Tejani, and Tae-Kyun Kim.
\newblock Latent regression forest: Structured estimation of 3d articulated
  hand posture.
\newblock In {\em Proceedings of the IEEE conference on computer vision and
  pattern recognition}, pages 3786--3793, 2014.

\bibitem{passive_ann_grasp}
Jonathan Tompson, Murphy Stein, Yann Lecun, and Ken Perlin.
\newblock Real-time continuous pose recovery of human hands using convolutional
  networks.
\newblock {\em ACM Transactions on Graphics (ToG)}, 33(5):169, 2014.

\bibitem{vollmer2017infrared}
Michael Vollmer and Klaus-Peter M{\"o}llmann.
\newblock {\em Infrared thermal imaging: fundamentals, research and
  applications}.
\newblock John Wiley \& Sons, 2017.

\bibitem{vision_grasp_tax_pred1}
Yezhou Yang, Cornelia Fermuller, Yi Li, and Yiannis Aloimonos.
\newblock Grasp type revisited: A modern perspective on a classical feature for
  vision.
\newblock In {\em The IEEE Conference on Computer Vision and Pattern
  Recognition (CVPR)}, June 2015.

\bibitem{grasp_from_contact}
Yuting Ye and C~Karen Liu.
\newblock Synthesis of detailed hand manipulations using contact sampling.
\newblock {\em ACM Transactions on Graphics (TOG)}, 31(4):41, 2012.

\bibitem{yuan2017bighand2}
Shanxin Yuan, Qi Ye, Bjorn Stenger, Siddhant Jain, and Tae-Kyun Kim.
\newblock Bighand2. 2m benchmark: Hand pose dataset and state of the art
  analysis.
\newblock In {\em Proceedings of the IEEE Conference on Computer Vision and
  Pattern Recognition}, pages 4866--4874, 2017.

\bibitem{colormap_optim}
Qian-Yi Zhou and Vladlen Koltun.
\newblock Color map optimization for 3d reconstruction with consumer depth
  cameras.
\newblock {\em ACM Transactions on Graphics (TOG)}, 33(4):155, 2014.

\bibitem{open3d}
Qian-Yi Zhou, Jaesik Park, and Vladlen Koltun.
\newblock {Open3D}: {A} modern library for {3D} data processing.
\newblock {\em arXiv:1801.09847}, 2018.

\bibitem{cyclegan}
Jun-Yan Zhu, Taesung Park, Phillip Isola, and Alexei~A Efros.
\newblock Unpaired image-to-image translation using cycle-consistent
  adversarial networks.
\newblock In {\em IEEE International Conference on Computer Vision}, 2017.

\end{thebibliography}
}
\end{document}